\documentclass[conference]{IEEEtran}
\IEEEoverridecommandlockouts
\usepackage{cite}
\usepackage{amsmath,amssymb,amsfonts}
\usepackage{graphicx}
\usepackage{textcomp}
\usepackage{xcolor}
\usepackage{tikz}
\usepackage{float}
\usepackage{booktabs}
\usepackage{multirow}
\usepackage{pgfplots}
\usepackage{comment}
\usepackage{subfigure}
\usepackage{hyperref}
\usepackage{url}

\usetikzlibrary{shapes}
\usetikzlibrary{arrows}
\def\BibTeX{{\rm B\kern-.05em{\sc i\kern-.025em b}\kern-.08em
    T\kern-.1667em\lower.7ex\hbox{E}\kern-.125emX}}
\begin{document}

\title{Combating the Elsagate Phenomenon: Deep Learning Architectures for Disturbing Cartoons
{\footnotesize }
\thanks{
Corresponding author: sandra@ic.unicamp.br.
A. Ishikawa is funded by PIBIC/CNPq, FAEPEX (\#2555/18) and Movile. 
E. Bollis is funded by CAPES. 
S. Avila is partially funded by Google Research Awards for Latin America 2018, FAPESP (\#2017/16246-0) and FAEPEX (\#3125/17).
RECOD Lab. is partially supported by diverse projects and grants from FAPESP, CNPq, and CAPES. 
We gratefully acknowledge the support of NVIDIA Corporation with the donation of the Titan Xp GPUs used for this research.}
}

\author{\IEEEauthorblockN{Akari Ishikawa, Edson Bollis, Sandra Avila}
\IEEEauthorblockA{RECOD Lab., Institute of Computing, University of Campinas (UNICAMP), Brazil}
}

\maketitle

\begin{abstract}
Watching cartoons can be useful for children's intellectual, social and emotional development. However, the most popular video sharing platform today provides many videos with Elsagate content. Elsagate is a phenomenon that depicts childhood characters in disturbing circumstances (e.g., gore, toilet humor, drinking urine, stealing). Even with this threat easily available for children, there is no work in the literature addressing the problem. As the first to explore disturbing content in cartoons, we proceed from the most recent pornography detection literature applying deep convolutional neural networks combined with static and motion information of the video. Our solution is compatible with mobile platforms and achieved 92.6\% of accuracy. Our goal is not only to introduce the first solution but also to bring up the discussion around Elsagate.
\end{abstract}


\section{Introduction}

Children today are part of a digital generation that has grown up in a world surrounded by technology (e.g., smartphones, tablets, electronic toys). They spend most of their time on the Internet, usually watching cartoons. Few children do not recognize the YouTube logo even from afar. 
Any smartphone or tablet is enough to navigate through the endless recommendation list generated by YouTube. Eventually, by following the suggested cartoons, it is unavoidable that one stumbles upon the Elsagate phenomenon~\cite{subreddit}.

In the Elsagate videos, Disney characters, superheroes and other popular childhood characters are depicted in disturbing scenes such as stealing alcohol, hurting each other, drinking from toilets, eating poop, drinking urine, smearing feces on people's faces, sexual and violent situations. 

We claim that the Elsagate videos are a way of getting young children accustomed to sexual and disturbing content so pedophiles can groom them more easily. However, there is no reliable evidence on whether those claims are real, what the real motivation is, or even who are the people responsible for making these videos \cite{forbeselsagate, bbc, nytimes}.

Elsagate channels have existed since 2014 \cite{koreanJongAng2018Crude}. In 2017, the term Elsagate (composed of Elsa, a character from the 2013 Disney animated film Frozen, and -gate, a suffix for scandals) became a popular hashtag on Twitter as users called attention to the presence of such material on YouTube and YouTube Kids. On Reddit, an Elsagate subreddit (r/Elsagate) was created to discuss the phenomenon, attracting tens of thousands of users~\cite{subreddit}.

As far as we know, there is no previous work in the literature related to the Elsagate phenomenon. Also, despite the existence of good solutions towards pornography/violence detection, Elsagate videos are wildly different in several aspects. The situation is even direr not only due to the lack of solutions geared towards cartoons but also due to the similitude among them and non-sensitive cartoons. In other words, classifying a video as Elsagate or not is a challenge itself.

In this paper, we come up with solutions that take advantage of deep neural networks for disturbing Elsagate content detection in cartoons. In a nutshell, our contributions are three-fold: 1) we propose an end-to-end pipeline (features, neural network architecture, classification model) to detect Elsagate content in videos; 2) we evaluate several deep neural networks proposed for mobile platforms, and 3) we introduce the first Elsagate dataset, which comprises 285 hours (1,028,106 seconds) of 1,396 Elsagate and 1,898 non-sensitive videos.

We organize the remaining of this paper into five sections: In Section~\ref{sec:related-works}, we review the works most related to Elsagate content detection. In Section~\ref{sec:methodology}, we describe the proposed pipeline. In Section~\ref{sec:experimental-setup}, we introduce the Elsagate dataset built and the evaluation metrics. In the Section~\ref{sec:results-discussion}, we discuss the experimental results. In the Section~\ref{sec:conclusion}, we conclude the paper proposing a solution for disturbing cartoons detection and suggesting approaches for future works.

\section{Related Work}\label{sec:related-works}

To the best of our knowledge, there are very few works related to sensitive content in cartoons. The most related ones are the following four works. 

Alghowinem~\cite{alghowinem2018saferyoutube} proposed a multimodal approach to detect inappropriate content in videos from YouTube Kids. For that, one-second slices are  extracted for analysis and classification. Image frames, audio signal, transcribed text and their respective features (e.g., temporal robust features (TRoF), mel-frequency cepstral coefficient (MFCC), bag-of-words) are extracted from each slice. These features are then fed into individual classifiers, which are combined using a threshold-based decision strategy. According to Alghowinem, the paper acted as a proof of concept. 
But, the pilot experiment is performed on three videos which are not even cartoons. 

Kahlil et al. \cite{khalil2016violencecartoon} tackled the problem of violence detection in cartoons. They exploited color features from seven color channels (gray, red, green, blue, hue, saturation and value). The average dominant value, calculated from the color histograms, is used as a threshold-based classifier. As pointed by the authors, those features do not contain sufficient contextual information for content analysis. They performed the experiments on 504 clips (16,654 seconds), of which 112 have violent content. Their dataset is not publicly available.

Khan et al.~\cite{khan2018violentcartoons} also explored violence detection in cartoons.
They proposed a three-layered video classification framework: keyframe extraction, feature extraction using scale-invariant feature transform (SIFT), feature encoding using Fisher vector image representation and classification using spectral regression kernel discriminant analysis (SRKDA). They evaluated their approach on 100 videos, collected from various sources. The dataset (not publicly available) comprises~nearly 2~hours (7,100~seconds) of 52 violent and 48 non-violent~videos.

Papadamou et al.~\cite{papadamou2019disturbed} studied the Elsagate phenomenon\footnote{This paper was not available before our submission.} using the video's titles, tags, thumbnails and general statistics (e.g., views, likes, dislikes). They proposed to process each type of feature using a different technique and to apply a fully-connected layer to combine their outputs. 
Despite the 82.8\% accuracy achieved, their solution can be easily fooled by the uploaders since it does not take into account the video content (e.g., frames) and can disguise the features analyzed (e.g., title, thumbnail) to hide the sensitive content. Their dataset is not publicly available.

In the face of the related works, it is clear that there is a lack of research specifically for the Elsagate problem. As a matter of fact, we have in the literature plenty of solutions for sensitive content analysis but these works are focused on real-life videos with humans, regardless of the type of sensitive content (e.g., nudity~\cite{lopes_2009_a,lopes_2009_b}, pornography~\cite{perez2016neurocomputing,caetano_2016,moreira2019if}, child pornography~\cite{vitorino2018jvci}, violence~\cite{recod14mediaeval,moreira2017wacv,peixoto2018}). 

\section{Methodology}\label{sec:methodology}

Our methodology is directly inspired by the work of Perez et al. \cite{perez2016neurocomputing}, a deep learning-based solution for video pornography detection. The main reason is their flexible pipeline that allows modifications in any of its steps, letting us approach the Elsagate problem in different ways. Fig.~\ref{fig:pipeline} depicts a flowchart overview of the proposed method.

Although Perez et al. obtained their best results combining raw frames and optical flows, for the sake of efficiency we opted to combine raw frames and MPEG motion vectors, which are computationally very cheap. In this paper, we aim an effective solution for mobile~platforms. Thus, we attempt to answer the following questions: 1) Is it better transferring knowledge (features) from a related dataset (e.g., Pornography) or an unrelated dataset (e.g., ImageNet)?, 2) Which deep learning architecture offers the best Elsagate classification performance regarding a mobile platform?

In the next subsections, we detail our pipeline. In Section~\ref{sec:features}, we present the static/motion features that we used as input to the deep neural networks. In Section~\ref{sec:dla}, we overview the deep learning architectures that we evaluated: GoogLeNet~\cite{szegedy2015cvpr}, SqueezeNet~\cite{iandola2016squeezenet}, MobileNetV2~\cite{sandler2018mobilenetv2}, and NASNet~\cite{zoph2018nasnet}. Finally, in Section~\ref{sec:late-fusion}, we describe the fusion strategy adopted in this work.

\tikzstyle{blue} = [rectangle, minimum width=1cm, minimum height=1cm, text width = 2cm,text centered, draw=blue!80, thick, fill=blue!30]
\tikzstyle{red} = [rectangle, rounded corners, minimum width=1cm, minimum height=1cm,text width = 2.2cm, text centered, draw=red!80, thick, fill=red!30]
\tikzstyle{green} = [rectangle, rounded corners, minimum width=0.5cm, minimum height=1cm,text width = 2.2cm,text centered, draw=green!80, thick, fill=green!30]
\tikzstyle{arrow} = [thick,->,>=stealth]
\begin{figure}[t]
\begin{center}
\begin{tikzpicture}[node distance=2cm]
\node (videos) [blue] {Videos};
\node (frames) [red, right of=videos, xshift=1cm, yshift=0.7cm] {Frames};
\node (motion) [red, right of=videos, xshift=1cm, yshift=-0.7cm] {Motion Vectors};
\node (cnn) [blue, right of=motion, xshift=1cm, yshift=0.7cm] {Deep Learning Architecture};
\node (pooling) [blue, below of=cnn, yshift=-0.4cm] {Pooling};
\node (latefusion) [blue, left of=pooling, xshift=-1cm] {Late Fusion};

\node (class) [green, left of=latefusion, xshift=-1cm] {Classification};

\draw [arrow] (cnn) -- (pooling);
\draw [arrow] (videos) -- (frames);
\draw [arrow] (frames) -- (cnn);
\draw [arrow] (videos) -- (motion);
\draw [arrow] (motion) -- (cnn);
\draw [arrow] (pooling) -- (latefusion);
\draw [arrow] (latefusion) -- (class);

\end{tikzpicture}
\caption{Overview of the proposed method for detecting Elsagate content.} 
\label{fig:pipeline}
\end{center}
\end{figure}
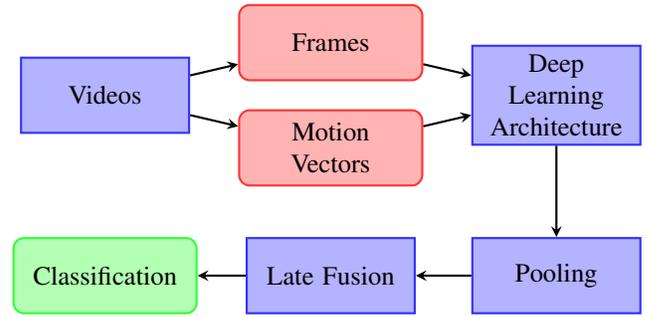

\subsection{Features}\label{sec:features}

\subsubsection{Static Information}

Deep neural networks have been successfully used to directly process raw data as input~\cite{krizhevsky2012imagenet,szegedy2015cvpr,sandler2018mobilenetv2}. To feed the video static information to the deep learning architectures, we extracted the raw frames from the videos using a frame sampling of one frame per second (1~fps). We then centered the frames, resized to 224$\times$224 pixels (the input size of GoogLeNet, SqueezeNet, MobileNetV2, and NASNet) maintaining the aspect ratio, and cropped the remaining pixels in the largest dimension.

\subsubsection{Motion Information}

Incorporating motion information into local descriptors and deep neural networks leads to more effective sensitive videos classifiers \cite{moreira2016fsi,moreira2017wacv,perez2016neurocomputing}. In this paper, we aim to develop deep learning-based approaches for automatically extracting discriminative space-temporal information for filtering Elsagate content, with a good compromise between effectiveness and efficiency. For that, we opt to extract MPEG motion vectors~\cite{motion-vector}, as suggested by Perez et al.~\cite{perez2016neurocomputing}.

Motion vectors can be decoded directly from standard video compressed files with very low computational cost. The decoding process includes many subprocesses such as motion compensation, inverse discrete cosine transform, and variable length decoding. The key process in decoding is the motion compensation by inter-frame prediction. In the motion compensation, a motion vector, which is a translation offset from the reference frame to the target frame, represents the movement of the macroblock (small regions in each frame). 

\begin{figure}[h]
\begin{center}
\subfigure{%
\includegraphics[height=1in]{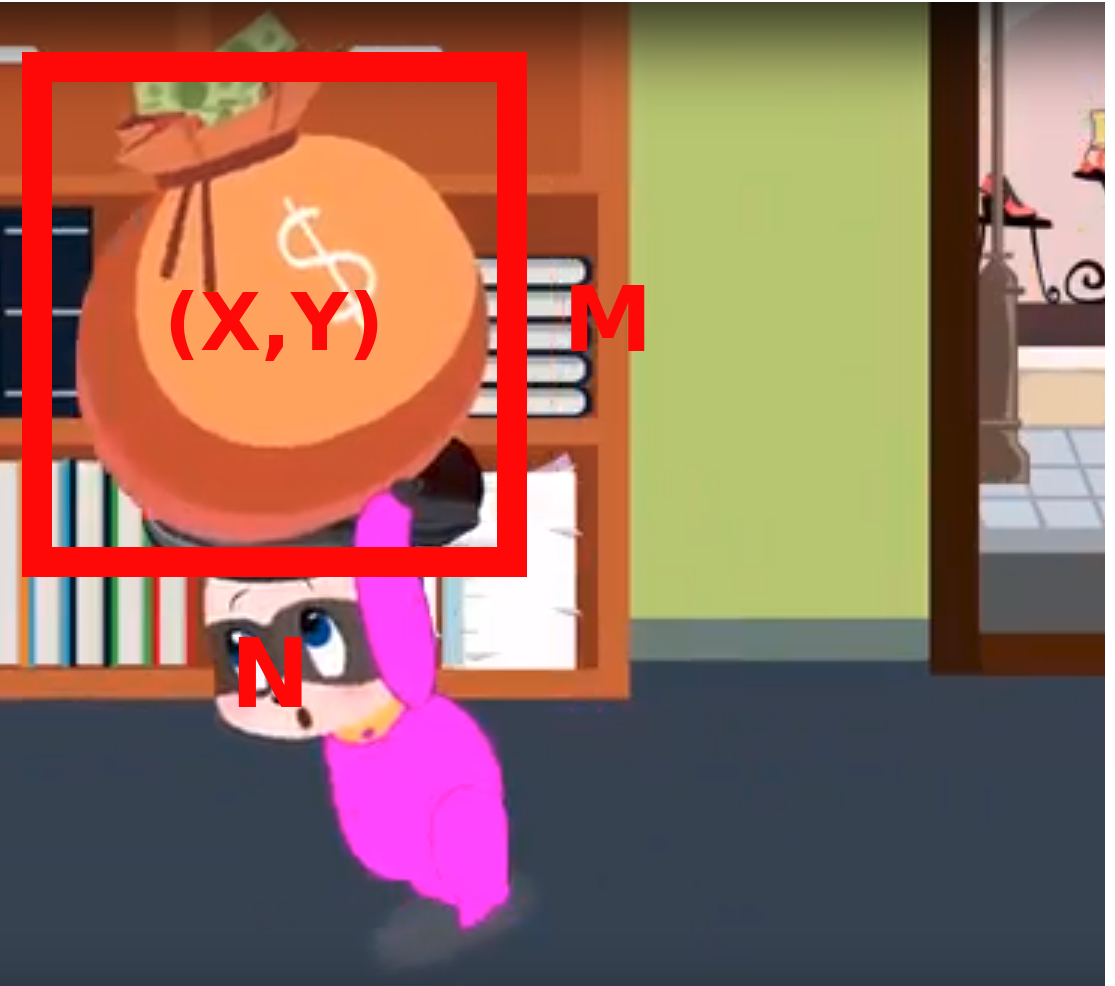}}%
\qquad
\subfigure{%
\includegraphics[height=1in]{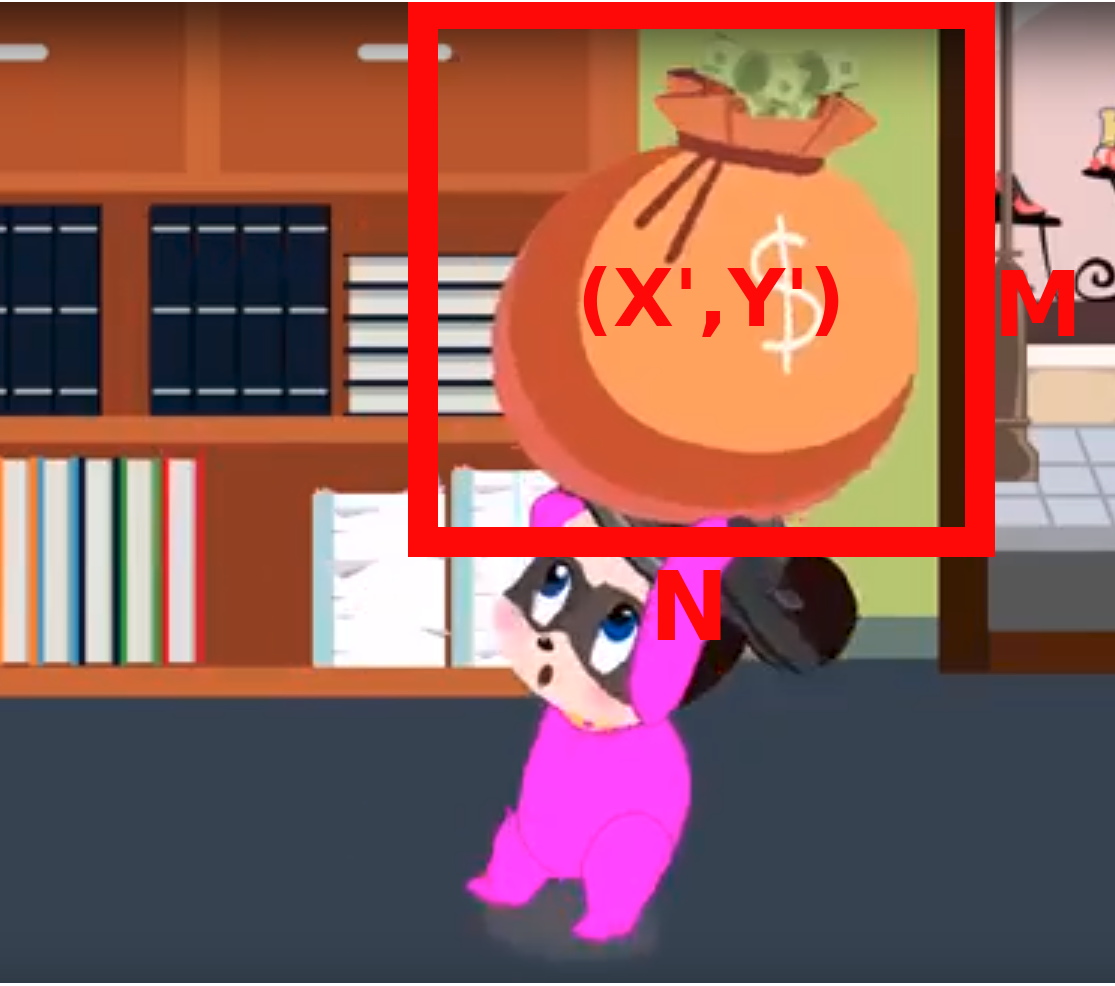}}%
\caption{Example of macroblock and its respective motion vector in the reference frame $(X,Y)$ and in the current frame $(X', Y')$ \cite{perez2016neurocomputing}. }
\label{fig:motionvector}%
\end{center}
\end{figure}

We calculate the MPEG motion vectors that map the motion between the reference and the current frame, using macroblocks of size $M\times N$ (usually $M$ and $N$ equals to 8 or 16).
Fig.~\ref{fig:motionvector} depicts a macroblock and its respective motion vector in the reference and the current~frame. 

\subsection{Deep Learning Architectures}\label{sec:dla}

Since the success of the ImageNet challenge~\cite{krizhevsky2012imagenet}, deep learning-based methods have drawn a tremendous amount of attention from academia, industry, and media. Subsequent works show that deep neural networks achieve the state-of-the-art results on many real-world tasks~\cite{LeCun2015deep}. In this paper, we evaluate four different deep learning architectures (DLAs) on Elsagate classification problem. 

\subsubsection{GoogLeNet}

GoogLeNet \cite{szegedy2015cvpr} won the 2014 ImageNet Challenge, with an error rate of 6.7\%, an improvement of 55\% compared to the AlexNet of Krizhevsky et al.~\cite{krizhevsky2012imagenet}. 

The network architecture is quite different from previous ones (VGGNet, ZFNet, and AlexNet). It performs convolution on input with three different sizes of filters (1$\times$1, 3$\times$3, 5$\times$5), and stacks all the outputs. This technique is called \textit{inception module}. Before its inception idea, most popular DNNs just stacked convolution layers deeper and deeper.

GoogLeNet (or Inception v1) has nine inception modules stacked linearly, 22 layers deep, and 6.8 million parameters. Also, it uses a global average pooling at the end of the last inception module.

\subsubsection{SqueezeNet}

SqueezeNet~\cite{iandola2016squeezenet} is an architecture that achieves AlexNet-level accuracy with 50$\times$ fewer parameters. 

The building block of SqueezeNet is called \textit{fire module}, which contains two layers: a \textit{squeeze} layer (convolution layer of only 1$\times$1 filters) and an \textit{expand} layer (convolution layer with a mix of 1$\times$1 and 3$\times$3 filters). The squeeze layer and expand layer keep the same feature map size: while the former reduces the depth to a smaller number, the latter increases it.

SqueezeNet contains eight fire modules, with a variable number of convolution filters, and uses 1.2 million parameters.  

Its last layer is a fully-connected layer with 1000 neurons and a softmax activation. Differently from other architectures evaluated in this paper, where we used the last (or previous ones) as the feature vector, none of the SqueezeNet's layers were suitable for our purpose. Either because of its size (for example, the layer immediately before the last one has a 512$\times$13$\times$13 dimension), either due to its sparsity. Thus, to make use of SqueezeNet, we added another fully-connected layer with two classes after the already existing one and considered the 1000 neurons layer as the feature vector. 

\subsubsection{MobileNetV2}

MobileNetV2~\cite{sandler2018mobilenetv2} is an architecture developed to mobile platforms. It uses separable convolutions to reduce the number of parameters: depthwise separable convolutions and bottleneck blocks. In depthwise convolution~\cite{sifre2014rigid}, convolution is performed independently for each of the input channels. This significantly reduces the computational cost by omitting convolution in the channel domain. Bottlenecks~\cite{gholami2018squeezenext} reduce the amount of data that flows through the network. In the bottleneck block, the output of each block is a bottleneck.

The architecture of MobileNetV2 contains the initial fully convolution layer with 32 filters, followed by 19 residual bottleneck layers, and uses 2.3 million parameters.

\subsubsection{NASNet}

NASNet~\cite{zoph2018nasnet} architecture is constructed using the Neural Architecture Search (NAS) framework. The goal of NAS is to use data-driven for constructing the network architecture. Szegedy et al.~\cite{szegedy2017inceptionv4} showed that a complex combination of filters in a `cell' can significantly improve results. The NAS framework defines the construction of such a cell as an optimization process and then stacks the multiple copies of the best cell to construct a large network.

In this paper, we use the NASNet-A (4 @ 1056), where 4 indicate the number of cell repeats and 1056 the number of filters in the penultimate layer of the network. It uses 5.3~million parameters.

\subsection{Late Fusion}\label{sec:late-fusion}

In this fusion scheme, each information is processed by a separate decision-making approach (e.g., support vector machine (SVM) classifier), generating independent classification scores that are combined on a single score for the final classification. Similarly to Perez et al.~\cite{perez2016neurocomputing}, we employ late fusion taking the mean of the probabilities from static and motion information, making a more precise classification. 

\section{Experimental Setup}\label{sec:experimental-setup}

We describe here the general experimental setup, such as the first Elsagate dataset and the metrics we use to assess the performance of the Elsagate classifiers.  All material related to this paper (dataset, code, models) is available at \url{https://github.com/akariueda/DLAforElsagate}.

To evaluate the results of our experiments, we apply a 1$\times$2-fold cross-validation protocol. It consists of randomly splitting the dataset into two folds. Then, we switched training and testing sets and, consequently, we conducted two analyses for every model.

\subsection{Elsagate Dataset}

We introduce the first publicly available Elsagate video dataset. It comprises 285 hours (1,028,106 seconds) of 1,396 Elsagate and 1,898 non-sensitive videos. To put the number in perspective, the largest sensitive video dataset (Pornography-2k dataset~\cite{moreira2016fsi}) contains 140 hours. It is worth mentioning the Elsagate dataset is composed of cartoons only. 

Concerning the Elsagate class, we downloaded videos from YouTube channels reported by Reddit users in the thread ``What is Elsagate?'' \cite{subreddit}. With respect to non-sensitive content, we collected videos from official YouTube channels (e.g., Cartoon Network, Disney Channel).

\begin{figure}[t]
\begin{center}
  \includegraphics[height=0.0775\textheight]{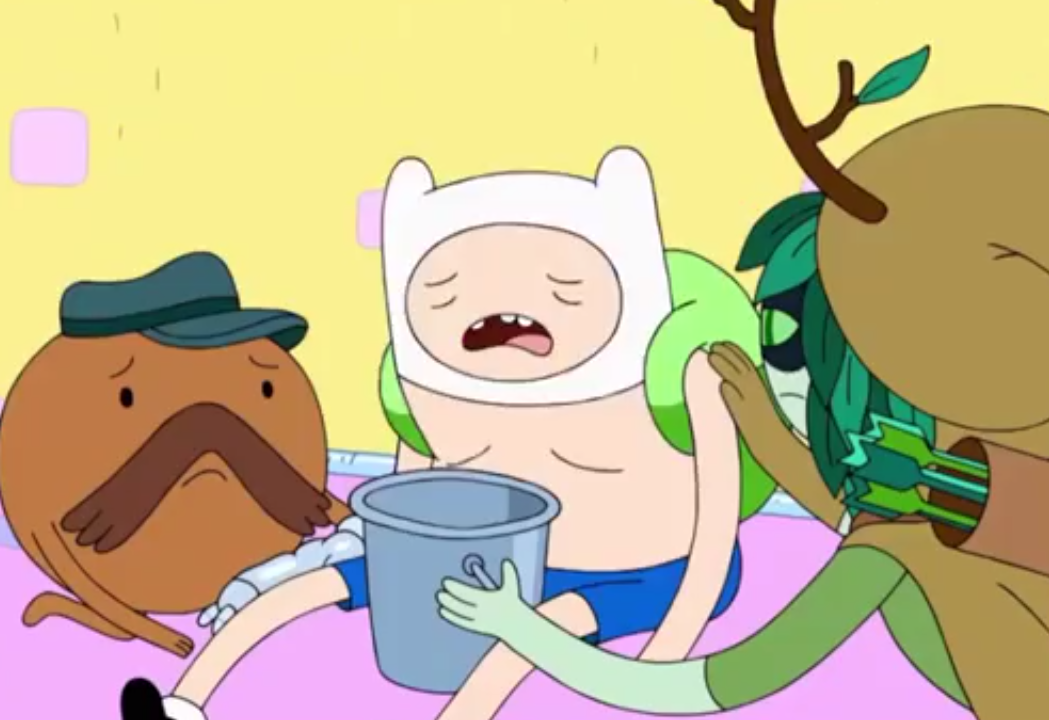}
  \includegraphics[height=0.0775\textheight]{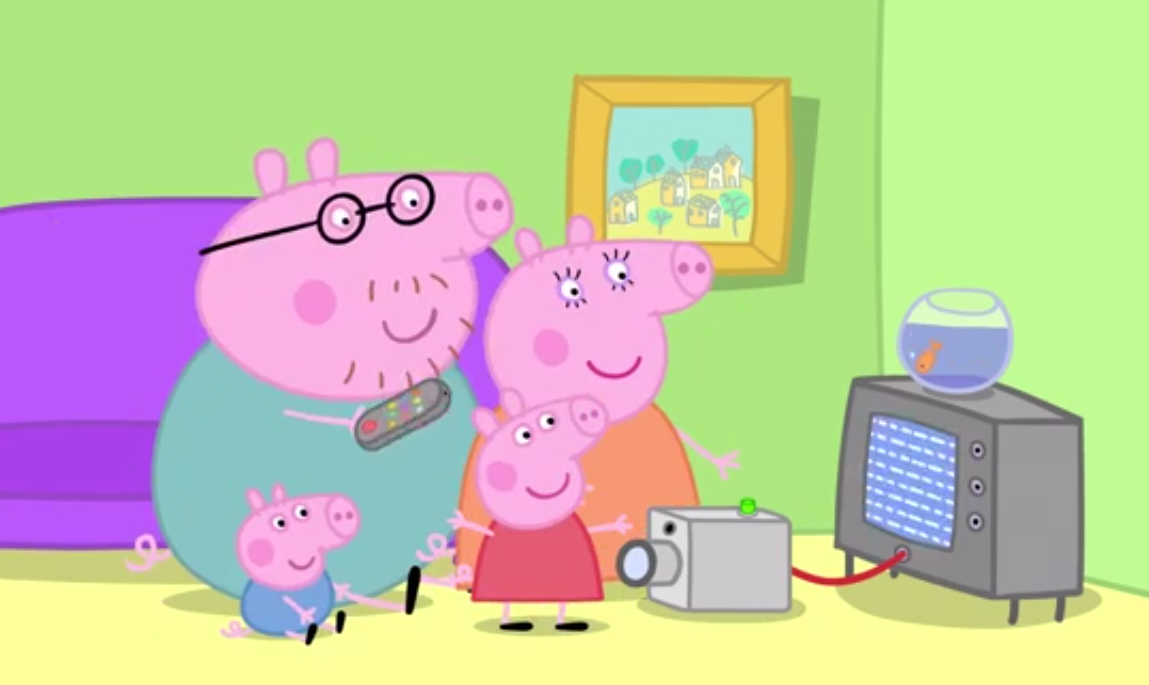}
  \includegraphics[height=0.0775\textheight]{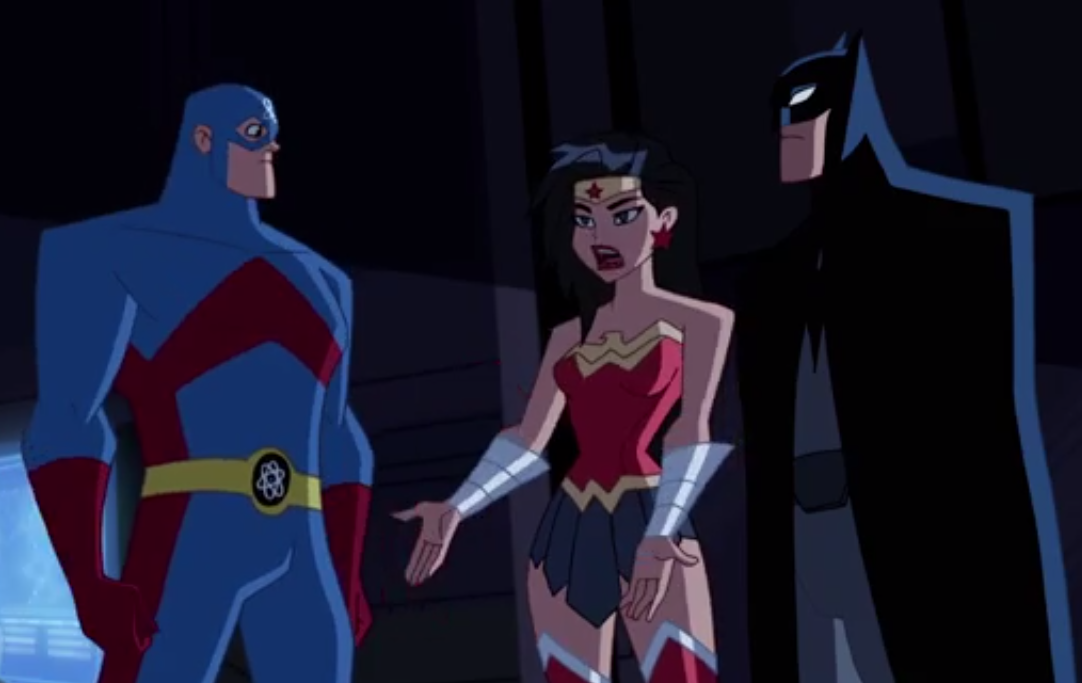}\\ \vspace{0.1cm}
  \footnotesize{(a) Non-sensitive}\\ \vspace{0.2cm}
  \includegraphics[height=0.077\textheight]{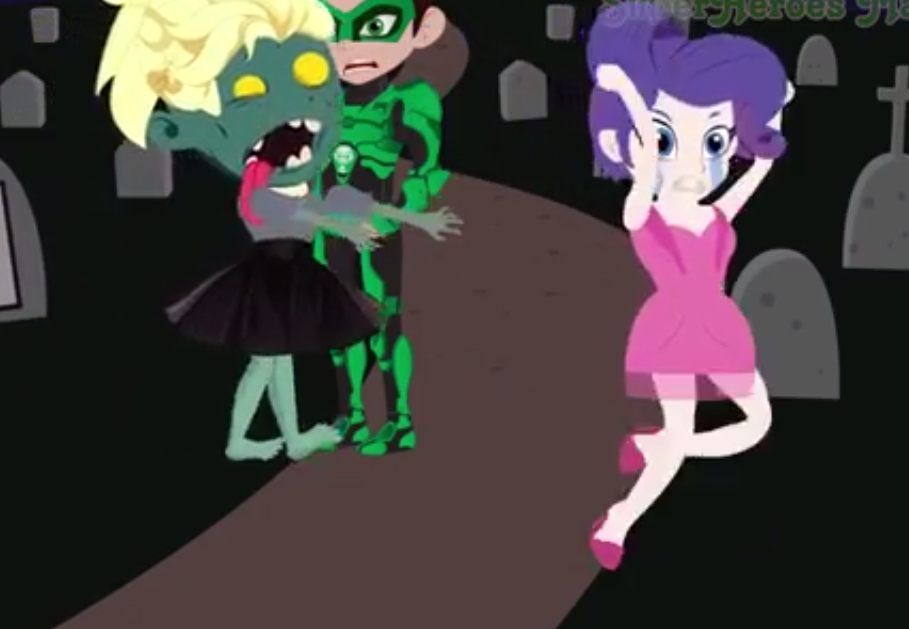}
  \includegraphics[height=0.077\textheight]{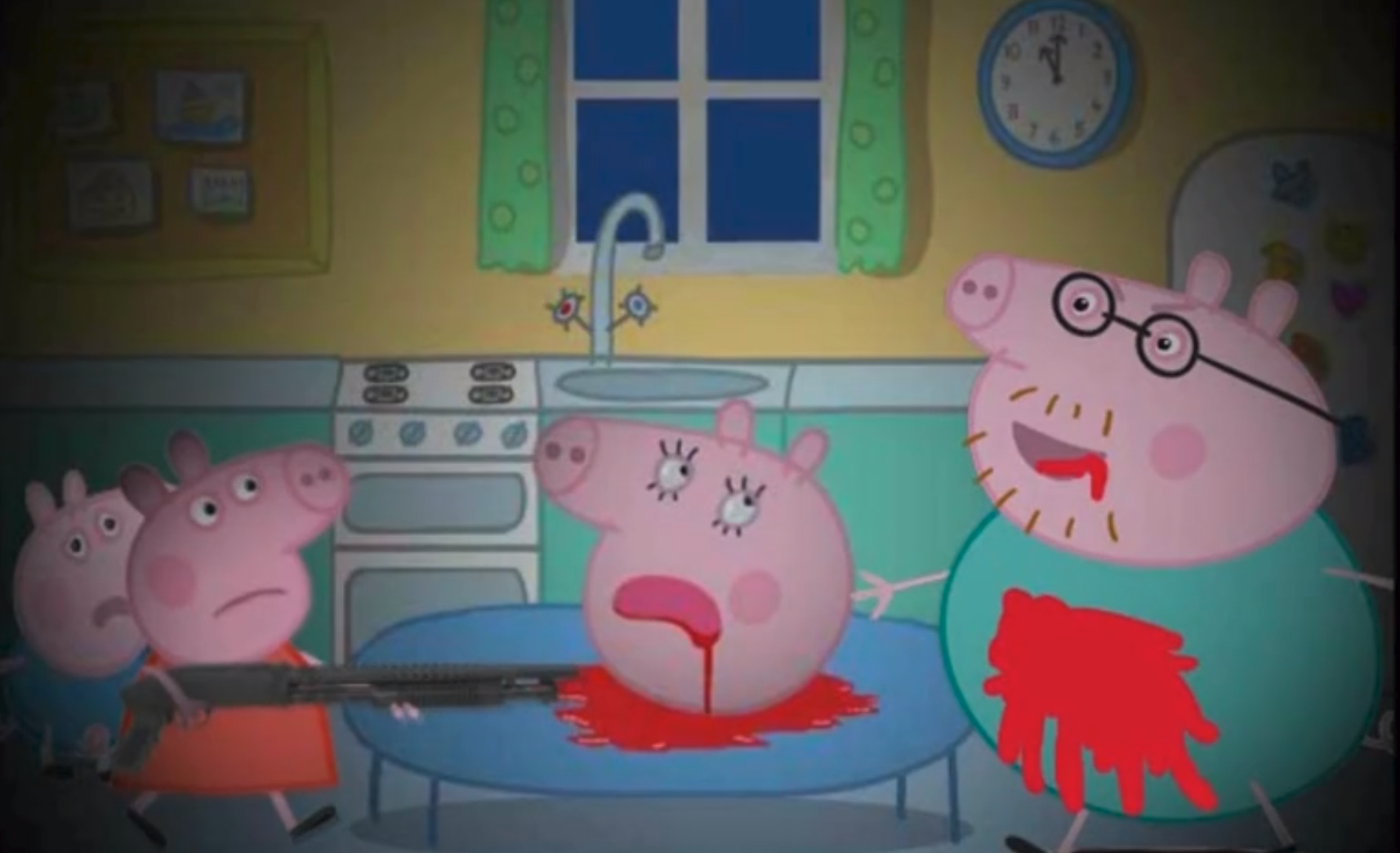}
  \includegraphics[height=0.077\textheight]{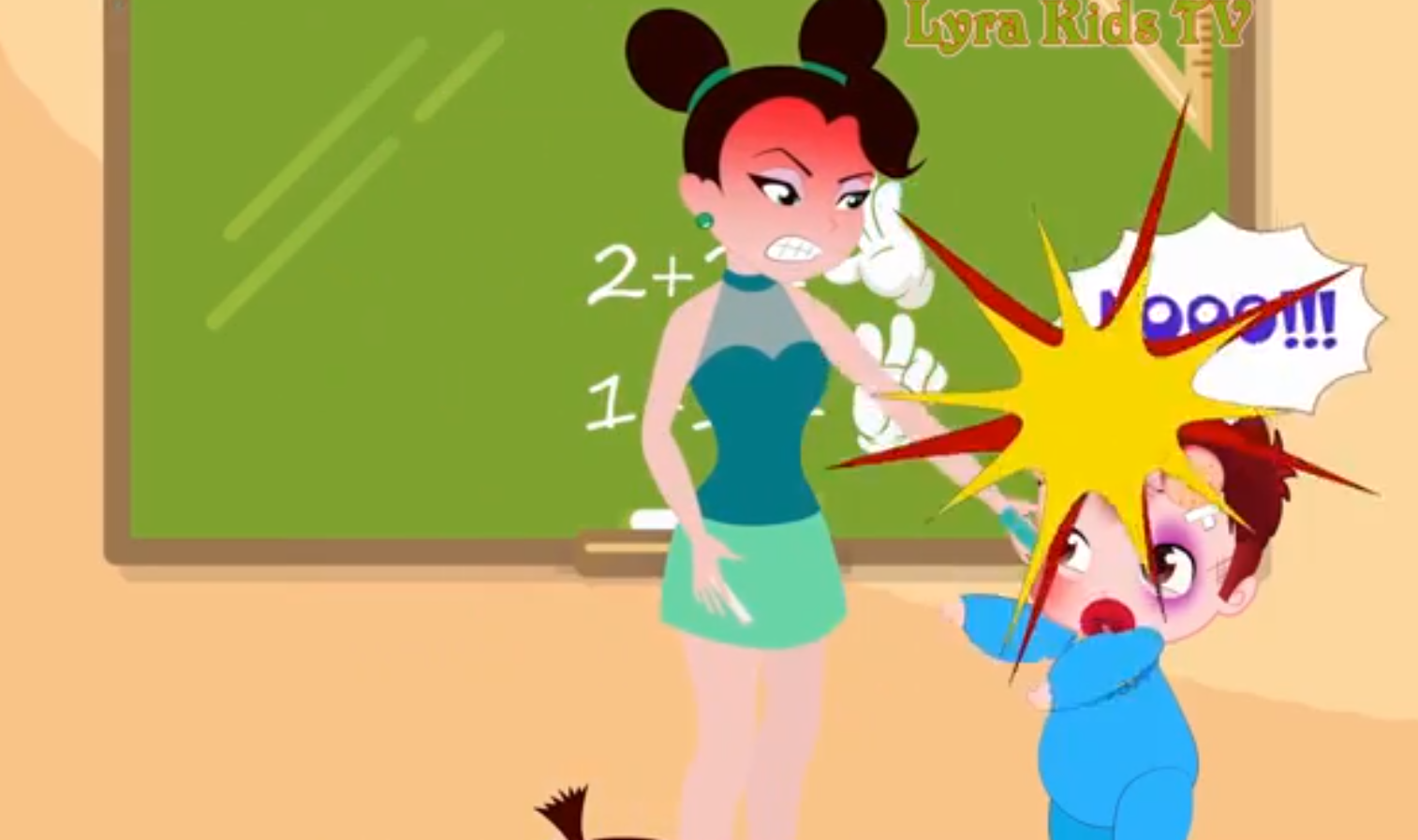}\\
  \vspace{0.1cm}
  \includegraphics[height=0.074\textheight]{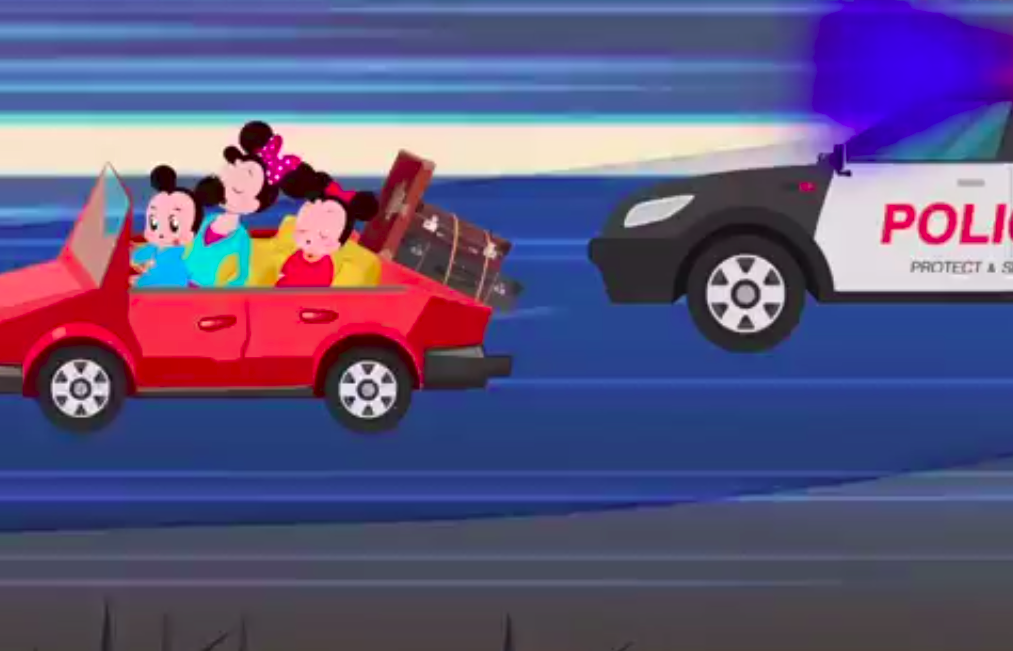}
  \includegraphics[height=0.074\textheight]{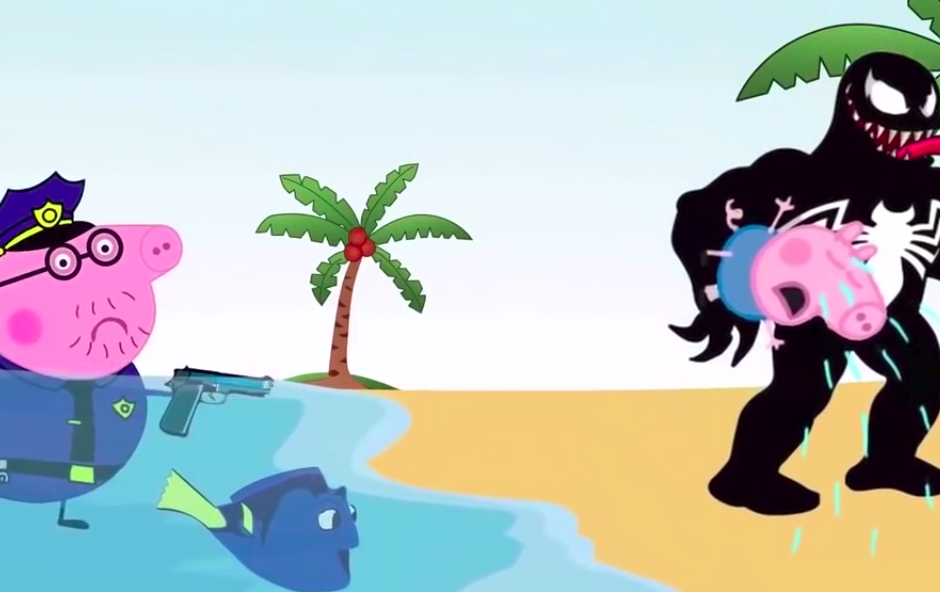}
  \includegraphics[height=0.074\textheight]{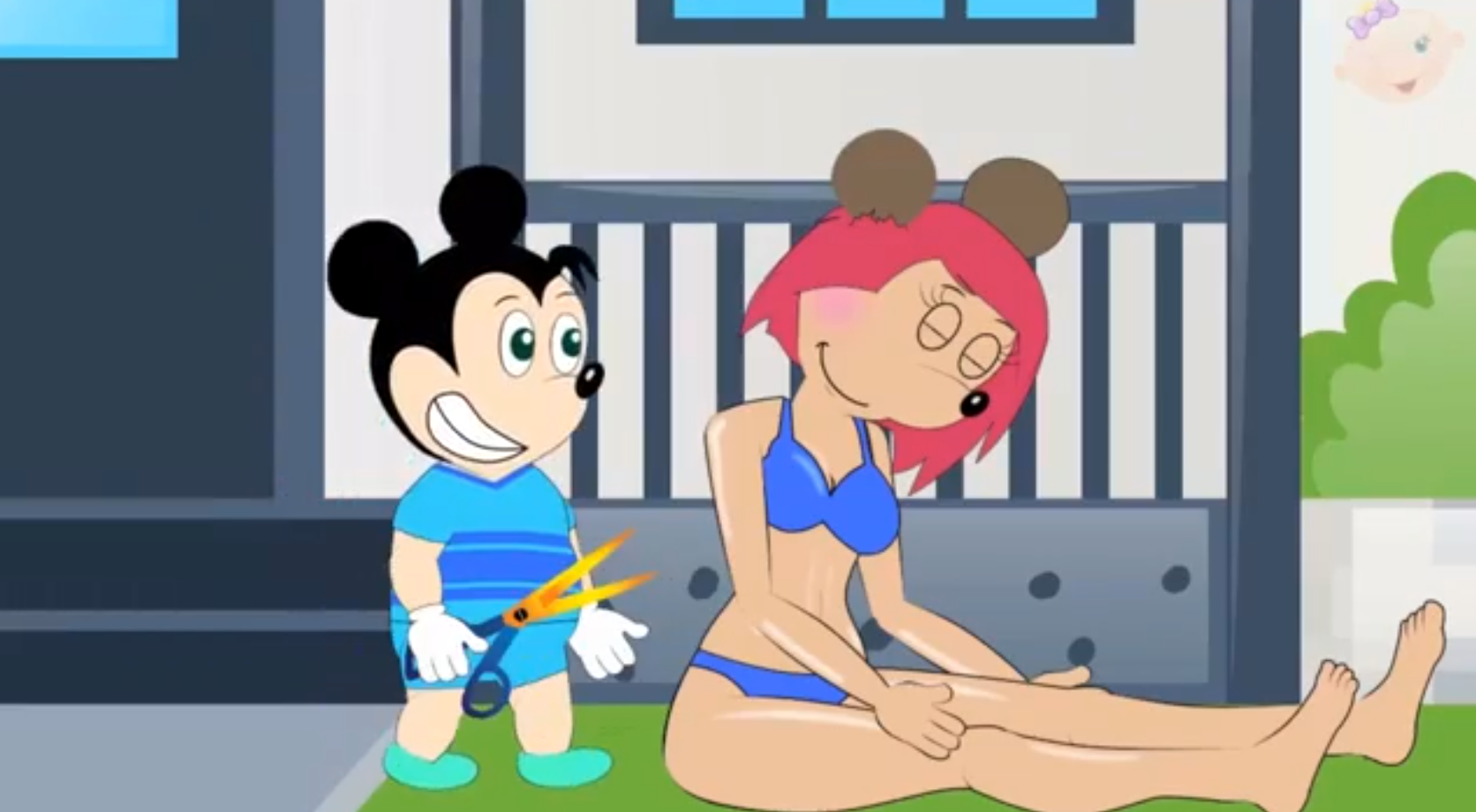}\\ \vspace{0.1cm}
  \footnotesize{(b) Elsagate}\\ 
\caption{Example frames from the Elsagate dataset.} \label{fig:database}
\end{center}
\end{figure}

On February 2018, we gathered a training/validation set with 1,567 non-sensitive and 1,118 Elsagate videos. On September 2018, we collected as a test set 331 non-sensitive and 278 Elsagate videos, totaling 1,898 non-sensitive and 1,396 Elsagates. The period between the two data gathering was purposeful to evaluate our model with new videos that could appear. The classes are imbalanced to offer a representative dataset. Fig.~\ref{fig:database} depicts some frames from the dataset. 

The Elsagate dataset is available free of charge to the scientific community but, due to the potential legal liabilities of distributing sensitive/copyrighted material, the request must be formal and a responsibility term must be~signed.

\subsection{Evaluation Metrics}\label{sec:evaluation-metrics}

To assess the performance of the Elsagate classifiers, we report the normalized accuracy and F\textsubscript{2}-measure in all experiments.

\begin{itemize}
\item Normalized Accuracy (ACC): measures the classifier's success rate, independently from the classes' labels. Mathematically, it is expressed as:
\[
ACC = (TPR+TNR)/2
\]

where $TPR$ is the True Positive Rate, and $TNR$ is the True Negative Rate.

\item F\textsubscript{2}-measure (F\textsubscript{2}): it is the harmonic mean of precision and recall where recall has double the precision's weight (\(\beta = 2\)). In this work, the F\textsubscript{2}-measure is crucial since the false negatives are unacceptable as it means the sensitive material passed the filtering. It is considered less harmful to wrongly deny the access to non-sensitive material than to expose sensitive content by mistake. The F\textsubscript{2}-measure can be defined as:
\begin{equation*}F\textsubscript{$ \beta $} = ( 1 + \beta^{2} ) \times \frac{\textit{precision} \times \textit{recall}}{\beta^{2} \times \textit{precision} + \textit{recall}}\end{equation*}
where the parameter $\beta$ denotes the importance of recall in relation to precision.
\end{itemize}

\section{Results and Discussion}\label{sec:results-discussion}

In this section, we evaluate the performance of different strategies on the Elsagate dataset.

\subsection{Transfer Learning}

\textit{Is it better transferring from a related dataset (e.g., Pornography) or an unrelated dataset (e.g., ImageNet)?} For this analysis, we used the GoogLeNet model, pre-trained for pornography in real-life (with humans) videos. Perez et al.~\cite{perez2016neurocomputing} kindly provided the weights trained for pornography.

Here, we aimed to evaluate the transfer learning exclusively. Note that in our pipeline (Fig.~\ref{fig:pipeline}), the deep neural network is used only as a feature extractor. In the classification step, we trained SVM models using our data. 

In Table~\ref{tab:pre-trainedmodel}, we observe the accuracy obtained with GoogLeNet pre-trained on ImageNet and Pornography videos. 
The network trained on ImageNet showed better results. For that reason, in the next experiments, we used pre-trained models on ImageNet.

\begin{table}[h]
\centering
\caption{Accuracy of the Perez et al. \cite{perez2016neurocomputing} model in Elsagate's videos.\vspace{-0.15cm}}
\begin{tabular}{lcc}
\toprule
Features                &   ImageNet (\%)      & Pornography (\%) \\ \midrule
Frames                  &   92.7         & 91.9   \\
Motion Vectors          &   91.3         & 92.3   \\
Late Fusion &   96.1         & 94.8   \\
\bottomrule
\end{tabular}
\label{tab:pre-trainedmodel}
\end{table}

\subsection{Mobile Deep Learning Architectures}

\begin{figure*}[]
\begin{center}
  \subfigure[Transfer Learning --- Accuracy]{\includegraphics[height=0.3\textwidth]{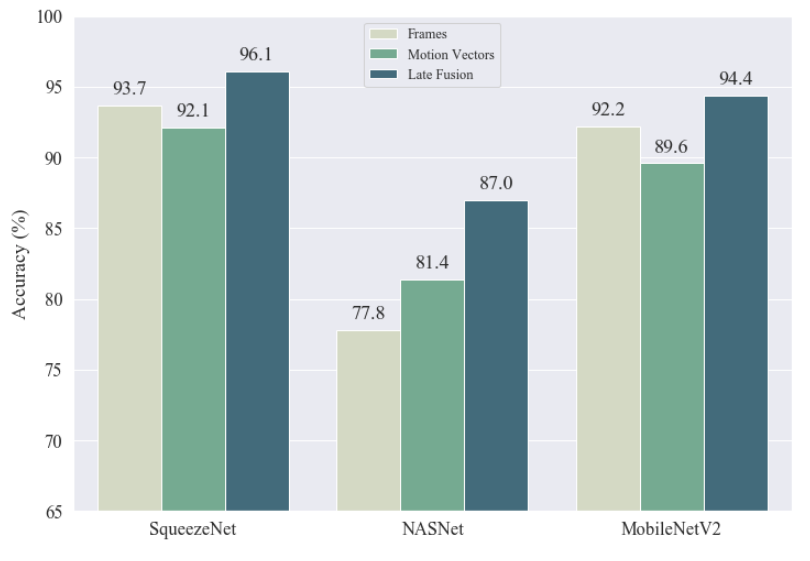}}
  \subfigure[Transfer Learning --- F2]{\includegraphics[height=0.3\textwidth]{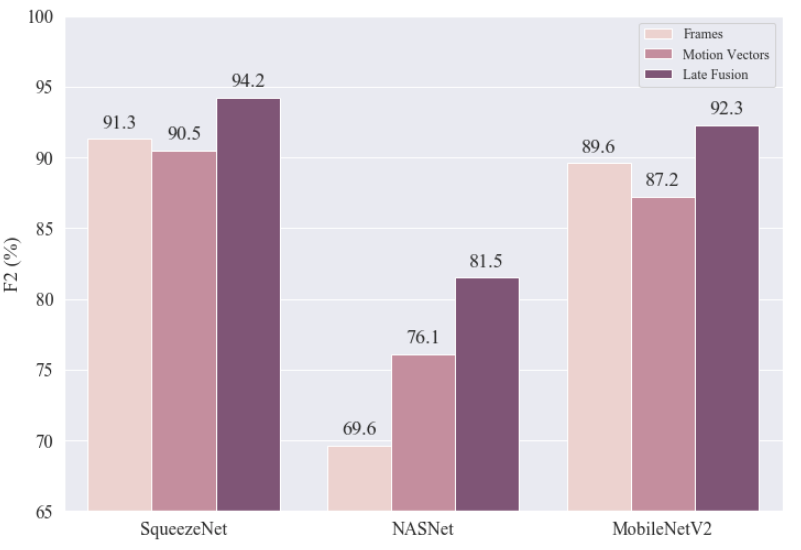}}\vspace{-0.1cm}
  \subfigure[Finetuning --- Accuracy]{\includegraphics[height=0.3\textwidth]{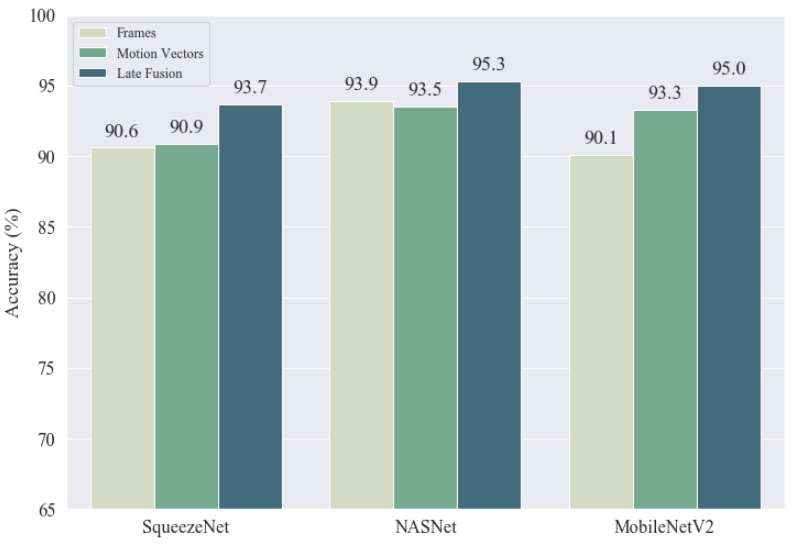}}
  \subfigure[Finetuning --- F2]{\includegraphics[height=0.3\textwidth]{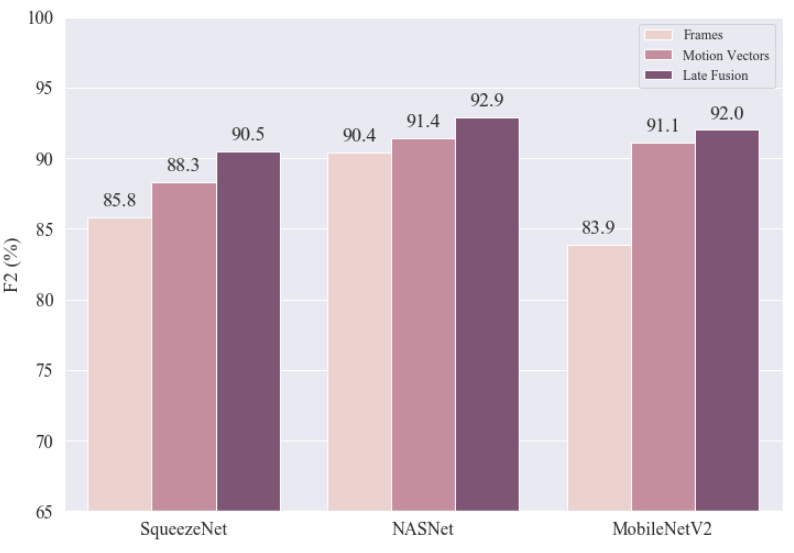}}
\caption{Accuracy and F\textsubscript{2} of SqueezeNet, NASNet and MobileNetV2 in transfer-learning and finetuning.\vspace{-0.2cm}} \label{fig:mobile-exps}
\end{center}
\end{figure*}

\textit{Which deep learning architecture offers the best Elsagate classification performance, regarding a mobile platform?} 
We reproduced the experiment of the previous section now evaluating SqueezeNet, NASNet, and MobileNetV2. Fig.~\ref{fig:mobile-exps} compares the accuracy and F\textsubscript{2}-measure of the three DLAs.

In the transfer learning (Fig.~\ref{fig:mobile-exps} (a) and (b)), SqueezeNet outperformed the other two architectures. But, it is worth mentioning that in preliminary experiments, using other SqueezeNet layers as feature vectors, it reported poor results. This indicates that SqueezeNet is not an adequate feature extractor for Elsagate content. The modifications we introduced to the architecture produced this curious outcome. 

Although transfer learning provided a reasonable suggestion about the DLAs feature extraction capacity, it still does not give us a sense of their learning capability. In the finetuning experiments, we initialized the DNNs with the transferred weights and then retrained all layers. Some experimentation involving freezing initial layers did not report good~results.

In Fig.~\ref{fig:mobile-exps} (c) and (d) we show the results for the finetuned models. Surprisingly, we noted that NASNet, which previously showed the worst performance, reached an accuracy of 95.3\% and F\textsubscript{2} of 92.9\%, surpassing both SqueezeNet and MobileNetV2 and competing closely to GoogLeNet, an architecture with the double of parameters. NASNet adaptation capability is comprehensible since it was developed through an architecture search method, as a joint of layers specialized for large datasets for classification (e.g., ImageNet) and detection tasks (e.g., COCO Object Detection).

In contrast with NASNet's improvements, we observed a reduction in SqueezeNet's results in all features (frames, motions and late fusion) and with MobileNetV2's frames. We hypothesize that the models that already report a fair performance (SqueezeNet achieved a 96.1\% accuracy in transfer learning) are likely to stumble upon a ``high accuracy barrier'' and end up unlearning the task.

\subsection{Testing}

Despite the interesting improvement in the NASNet-based model, we note that SqueezeNet's results are slightly higher. Since it is expected that finetuning a model brings better results, we downloaded a test set to help us decide upon this dilemma and define our final model.
As a decisive experiment, we also used the SVM models trained on the training set.

Although we had downloaded the test set around seven months later than the training set, a reliable model is supposed to be robust enough to deal with the unexpected. The results, shown in Table~\ref{tab:squeezenet-nasnet-test}, illustrate an impressive decline in SqueezeNet's performance. Thus, as our final~model, we chose NASNet transferred from ImageNet and finetuned to Elsagate.

\begin{table}[h]
\caption{SqueezeNet transferred model and NASNet finetuned model predicting the test set.\vspace{-0.15cm}}
\centering
\begin{tabular}{lcccc}
\toprule
                        & \multicolumn{2}{c}{Train Set} & \multicolumn{2}{c}{Test Set}    \\ 
                        & ACC (\%) & F2 (\%) & ACC (\%) & F2 (\%)\\ \midrule
SqueezeNet (Transfer)           & 96.1 & 94.2 & 62.0 & 37.8 \\
NASNet (Finetuned)           & 95.3  & 92.9 & 92.6  & 88.7 \\
\bottomrule
\end{tabular}
\label{tab:squeezenet-nasnet-test}
\end{table}

\subsection{Model Interpretation} 

At the same time that accuracy and F\textsubscript{2} are proper metrics to compare the architectures, they do not tell much about the model interpretability. Until now, we have only assessed the models regarding the number of Elsagate videos they were able to filter. Thus, an interesting object of study is to understand the model's behavior towards the videos and to speculate possible reasons that led to errors. Here, we analyze and discuss the false negatives and false positives  all the three architectures.

Although Elsagate is calling people's attention due to the bizarre topics depicted (e.g., gore or toilet humor), some softer topics such as nursery rhymes and counting numbers and colors are also considered Elsagate (accordingly to /r/Elsagate), leading to some controversy. Due to the subjectivity of the Elsagate content, we could say that the low-quality production of the video classifies it as an Elsagate, or also that the~characters expressions, sounds, and groanings may cause a disturbance. 

\begin{figure}[t]
\begin{center}
  \subfigure[Elsagate nursery rhyme]{    \includegraphics[height=0.135\textwidth]{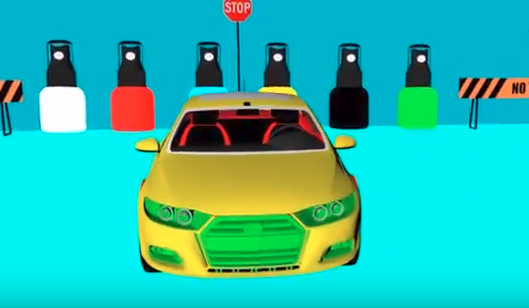}}
  \subfigure[Nursery rhyme]{ \includegraphics[height=0.135\textwidth]{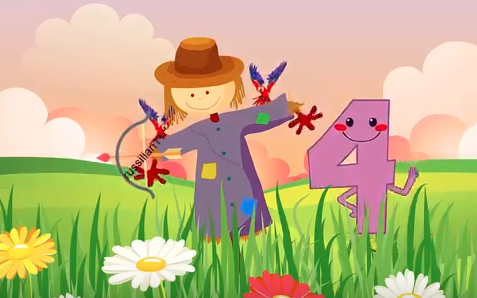}}
  \subfigure[False negative]{    \includegraphics[height=0.128\textwidth]{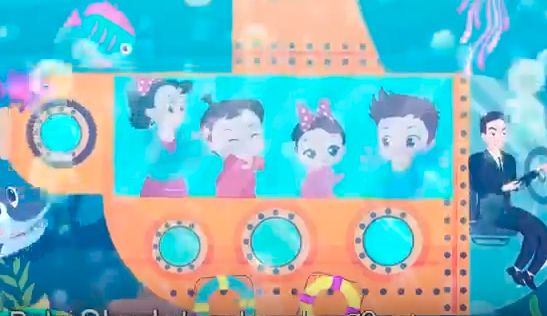}}
  \subfigure[False positive]{ \includegraphics[height=0.128\textwidth]{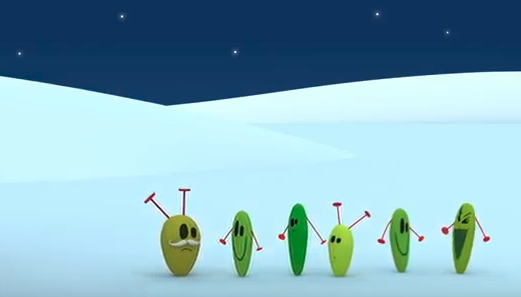}}
 
\caption{Sample frames of videos misclassified by all the architectures.} \label{fig:finetune-errors}
\end{center}
\end{figure}

Not surprisingly, most of the misclassified videos were nursery or counting rhymes. We also noted the presence of this kind of video in both classes, although the ones in the non-sensitive class have a much higher quality (Fig.~\ref{fig:finetune-errors} (a) and (b)). Therefore, regarding those videos, even if considered Elsagate, they are much less harmful than the average. So, we believe this model's behavior is acceptable.

Concerning the other false positives and negatives, in some videos the same Elsagate characters were not in a grotesque circumstance (Fig.~\ref{fig:finetune-errors} (c)). Also, in other videos, the appearance (e.g., color-palette, motion patterns) resembles Elsagate (Fig.~\ref{fig:finetune-errors}~(d)). Besides, we had many non-sensitive videos containing controversial scenes (e.g., fighting, naughtiness, accidents) that could be considered Elsagate content.

\section{Conclusion}\label{sec:conclusion}

In this paper, we brought up to the literature the discussion about the Elsagate phenomenon, and we proposed the first solution to solve the problem. Inspired by the most recent approaches in sensitive content detection, we evaluated popular DLAs such as GoogLeNet, SqueezeNet, NASNet and MobileNetV2 and both static and motion information of the video. Our experiments suggested that NASNet is an excellent feature extractor when applying transfer learning from ImageNet, followed by a finetuning to Elsagate. Even with the challenges imposed by the problem itself (e.g., the lack of a formal definition of Elsagate), we achieved a 92.6\% of accuracy. As future work, we intend to embed the solution in a mobile application and to propose a more deep annotation for studying the phenomenon itself. We hope this work stimulates the development of better disturbing content filtering solutions.

\bibliographystyle{IEEEtran}
\bibliography{refs}

\end{document}